\pdfoutput=1

\documentclass[11pt]{article}

\usepackage{ACL2023}

\usepackage{times}
\usepackage{latexsym}
\usepackage{graphicx}
\usepackage[algoruled,ruled,vlined,noend]{algorithm2e}

\usepackage{amsmath}
\usepackage{pgfplots}
\usepackage{graphicx}
\usepackage{enumitem}
\usepackage{tikz}
\usepackage{diagbox}
\usepackage{tkz-tab}
\usepackage{caption}
\usepackage{booktabs}
\usepackage{amssymb}
\usepackage{amsmath}
\usepackage{subcaption}
\usepackage{natbib}
\usepackage{blindtext}
\usepackage{url}

\usepackage{url}
\usepackage{hyperref}    
\usepackage{cleveref}
\SetAlFnt{\small}
\SetAlCapFnt{\small}
\SetAlCapNameFnt{\small}
\usepackage{ntheorem}

\usepackage{booktabs, tabularx}
\usepackage{tfrupee}  
\usepackage{caption}

\usepackage[T1]{fontenc}
\usepackage[utf8]{inputenc}

\usepackage{microtype}

\usepackage{inconsolata}

\title{Do GPTs Produce Less Literal Translations?}

\author{Vikas Raunak \qquad Arul Menezes \qquad Matt Post \qquad Hany Hassan Awadalla\\\\
Microsoft Azure AI \\
Redmond, Washington \\
\texttt{\{viraunak,arulm,mpost,hanyh\}@microsoft.com}}

\begin{document}
\maketitle
\begin{abstract}

Large Language Models (LLMs) such as GPT-3 have emerged as general-purpose language models capable of addressing many natural language generation or understanding tasks. On the task of Machine Translation (MT), multiple works have investigated few-shot prompting mechanisms to elicit better translations from LLMs. However, there has been relatively little investigation on how such translations differ \emph{qualitatively} from the translations generated by standard Neural Machine Translation (NMT) models. In this work, we investigate these differences in terms of the literalness of translations produced by the two systems.
Using literalness measures involving word alignment and monotonicity, we find that translations out of English (E$\rightarrow$X) from GPTs tend to be less literal, while exhibiting similar or better scores on MT quality metrics.
We demonstrate that this finding is borne out in human evaluations as well.
We then show that these differences are especially pronounced when translating sentences that contain idiomatic expressions.

\end{abstract}

\section{Introduction}
\label{sec:intro}

Despite training only on a language-modeling objective, with no \textit{explicit} supervision on aligned parallel data \cite{palm_data}, LLMs such as GPT-3 or PaLM \cite{gpt3, palm} achieve close to state-of-the-art translation performance under few-shot prompting \cite{mt_incontext_1, mt-gpt}.
Work investigating the output of these models has noted that the gains in performance are not visible when using older surface-based metrics such as BLEU \cite{papineni-etal-2002-bleu}, which typically show large losses against NMT systems.
This raises a question: How do these LLM translations differ \emph{qualitatively} from those of traditional NMT systems?

\begin{table}[t]
\begin{tabular}{|lp{2.25in}|}
\toprule
   source  & He survived by \colorbox{yellow}{the skin of his teeth}. \\
\midrule
   NMT  & Il a survécu par  \colorbox{pink}{la peau de ses dents}.  \\
   GPT-3 & Il a survécu \colorbox{lime}{de justesse}. \\
   \bottomrule
\end{tabular}
\caption{An example where GPT-3 produces a more natural (non-literal) translation of an English idiom. When word-aligning these sentences, the source word \textit{skin} remains unaligned for the GPT-3 translation.}%
\label{fig:example_page1}%
\vspace{-1.5em}
\end{table}

We explore this question using the property of translation \emph{literalness}.
Machine translation systems have long been noted for their tendency to produce overly-literal translations \cite{overly_literal}, and we have observed anecdotally that LLMs seem less susceptible to this problem (Table \ref{fig:example_page1}).
We investigate whether these observations can be validated quantitatively.
First, we use measures based on word alignment and monotonicity to quantify whether LLMs produce less literal translations than NMT systems, and ground these numbers in human evaluation (\S~\ref{sec:measures}).
Next, we look specifically at idioms, comparing how literally they are translated under both natural and synthetic data settings (\S~\ref{sec:idioms}).

Our investigations focus on the translation between English and German, Chinese, and Russian, three typologically diverse languages.
Our findings are summarized as follows: (1) We find that translations from two LLMs from the GPT series of LLMs are indeed generally less literal than those of their NMT counterparts when translating \textit{out} of English, and (2) that this is particularly true in the case of sentences with idiomatic expressions.

\section{Quantifying Translation Literalness}
\label{sec:measures}

\begin{table*}
\centering
\scalebox{0.70}{
\begin{tabular}{l|l|l}
\midrule
\textbf{System} & \textbf{Source}& \textbf{Translation}\\ \midrule
MS    & Time is running out for Iran nuclear deal, Germany says, & Die Zeit für das Atomabkommen mit dem Iran läuft ab, sagt Deutschland\\ 
GPT    & Time is running out for Iran nuclear deal, Germany says, & Deutschland sagt, die Zeit für das iranische Atomabkommen läuft ab. \\ \midrule
MS   & You're welcome, one moment please. & Sie sind willkommen, einen Moment bitte.\\ 
GPT    & You're welcome, one moment please. & Bitte sehr, einen Moment bitte.\\ \bottomrule
\end{tabular}}
\caption{Translation examples with different Non-Monotonicity (NM) and Unaligned Source Word (USW) scores for MS-Translator (lower) and text-davinci-003 translations (higher) from the WMT-22 En-De test set, for illustration.}
\label{tab:literalness_measurement_examples}
\vspace{-1.00em}
\end{table*}

We compare the state-of-the-art NMT systems against the most capable publicly-accessible GPT models (at the time of writing) across measures designed to capture differences in translation literalness.
We conduct both automatic metric-based as well as human evaluations. We explain the evaluation and experimental details below.

\paragraph{Datasets} We use the official WMT21 En-De, De-En, En-Ru and Ru-En News Translation test sets for evaluation \cite{wmt-2021-machine}.

\paragraph{Measures of Quality} We use COMET-QE\footnote{wmt20-comet-qe-da} \cite{comet} as the Quality Estimation (QE) measure \cite{qe_fomicheva} to quantify the fluency and adequacy of translations. Using QE as a metric presents the advantage that it precludes the presence of any reference bias, which has been shown to be detrimental in estimating the LLM output quality in related sequence transduction tasks \cite{goyal2022news}. On the other hand, COMET-QE as a metric suffers from an apparent blindness to copy errors (i.e., cases in which the model produces output in the source language) \cite{blind_spot}. To mitigate this, we apply a language identifier \cite{fastext-langid} on the translation output and set the translation to null if the translation language is the same as the source language. Therefore, we name this metric COMET-QE + LID.

\paragraph{Measures of Translation Literalness} There do not exist any known metrics with high correlation geared towards quantifying translation literalness. We propose and consider two automatic measures at the corpus-level:

\begin{enumerate}
    \item \textit{Unaligned Source Words (USW)}: Two translations with very similar fluency and adequacy could be differentiated in terms of their literalness by computing word to word alignment between the source and the translation, then measuring the number of source words left unaligned. 
    When controlled for quality, a less literal translation is likely to contain more unaligned source words (as suggested in Figure~\ref{fig:example_page1}). 
    \item \textit{Translation Non-Monotonicity (NM)}: Another measure of literalness is how closely the translation tracks the word order in the source. We use the non-monotonicity metric proposed in \citet{additive_interventions}, which computes the deviation from the diagonal in the word to word alignment as the non-monotonicity measure. This can also be interpreted as (normalized) alignment crossings, which has been shown to correlate with translation non-literalness \cite{literalness_measuring}.
\end{enumerate}
We use the multilingual-BERT-based awesome-aligner \cite{devlin-etal-2019-bert,awesome_aligner} to obtain the word to word alignments between the source and the translation. Table~\ref{tab:literalness_measurement_examples} presents an illustration of translations with different USW and NM scores\footnote{Metrics: \url{https://github.com/vyraun/literalness}}, obtained from different systems.

\paragraph{Systems Under Evaluation} We experiment with the below four systems (NMT and LLMs):
\begin{enumerate}
    \item WMT-21-SOTA: The Facebook multilingual system \cite{facebook_wmt21} won the WMT-21 News Translation task \cite{wmt-2021-machine}, and thereby represents the strongest NMT system on the WMT'21 test sets.
    \item Microsoft-Translator: MS-Translator is one of the strongest publicly available commercial NMT systems \cite{salted}.
    \item text-davinci-002: The text-davinci-002 model is an instruction fine-tuned model in the GPT family \cite{gpt3}. It represents one of the strongest publicly-accessible LLMs \cite{helm}.
    \item text-davinci-003: The text-davinci-003 model further improves upon text-davinci-002 for many tasks\footnote{LLMs: \url{https://beta.openai.com/docs/models/}} \cite{helm}. 
\end{enumerate}
For both the GPT models, we randomly select eight samples from the corresponding WMT-21 development set, and use these in the prompt as demonstrations for obtaining all translations from GPTs. 

\begin{figure}[ht!]
       \includegraphics[width=0.48\textwidth,height=0.3\textwidth]{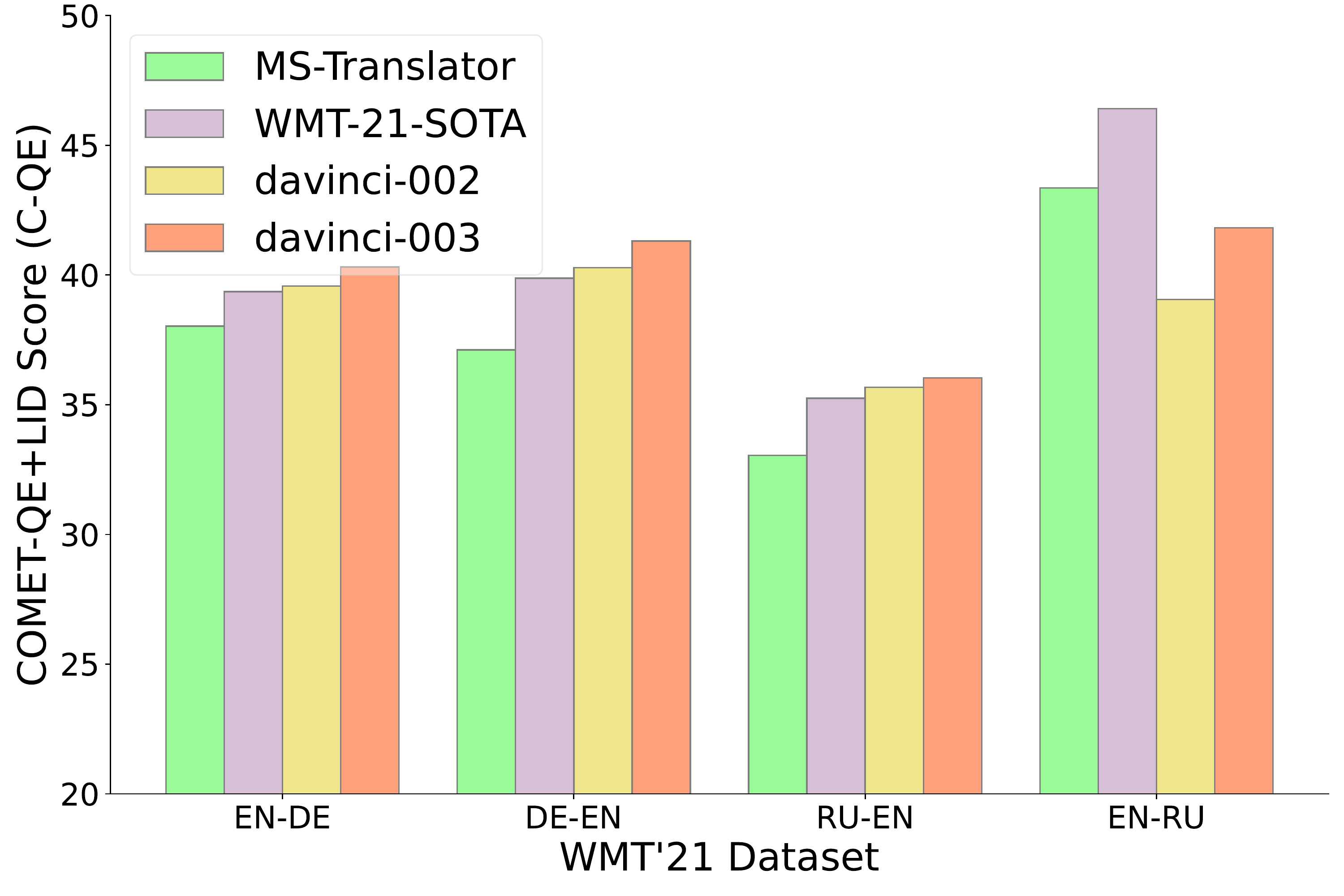}
       \includegraphics[width=0.48\textwidth,height=0.3\textwidth]{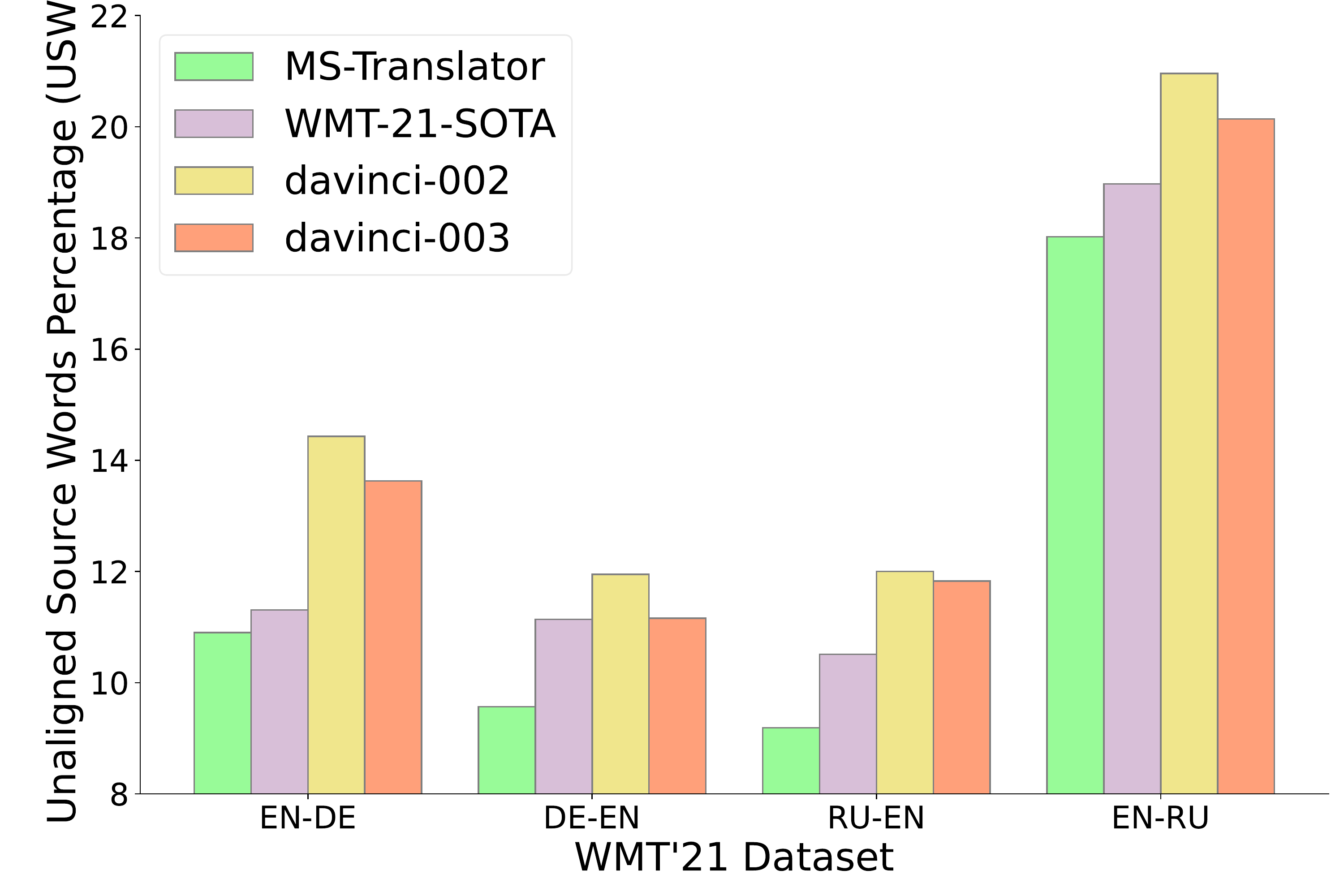}
       
\includegraphics[width=0.48\textwidth,height=0.3\textwidth]{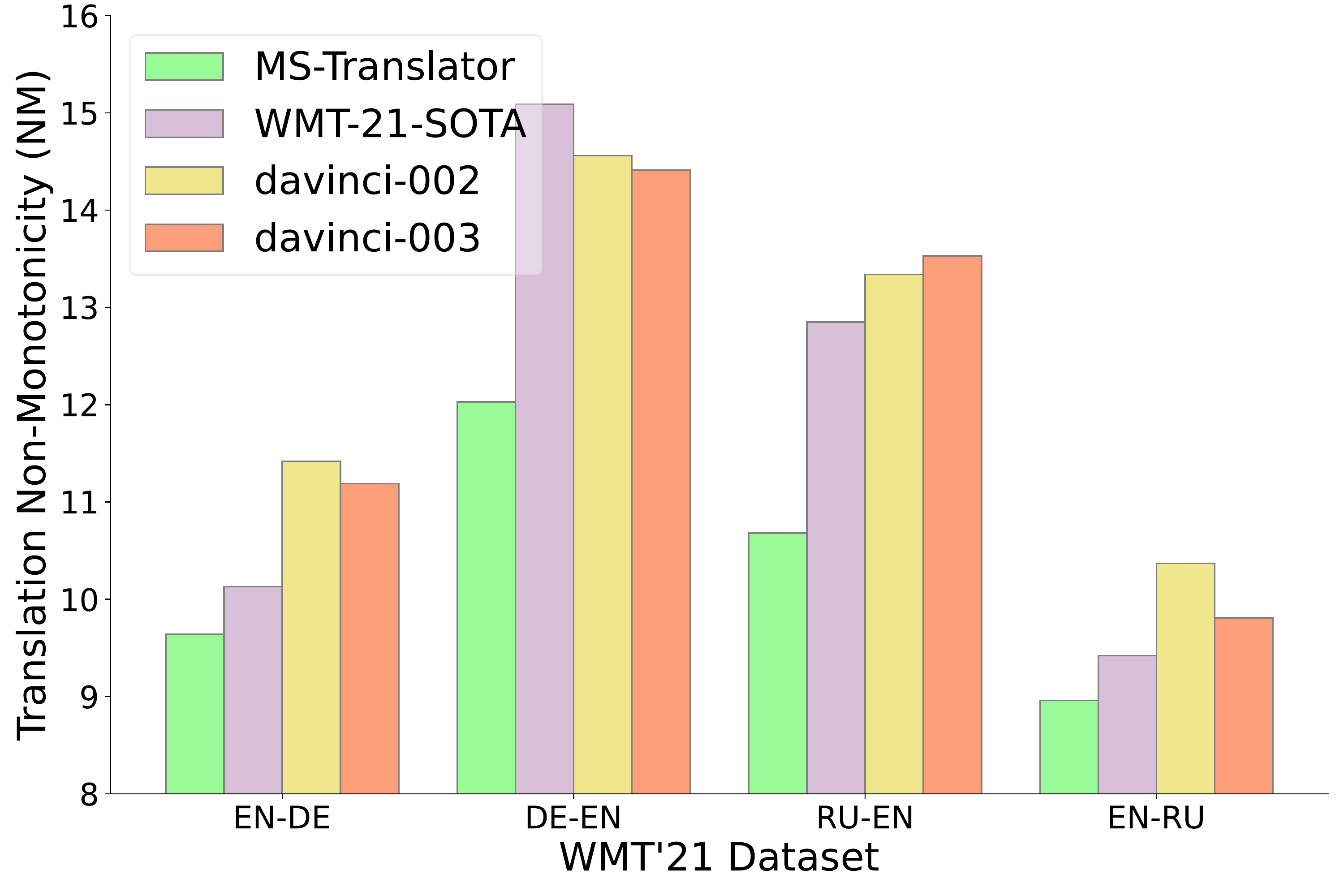}       
    \caption{Measurements: The NMT Systems and GPT models achieve similar COMET-QE+LID Scores (Top), there exists a significant gap in the number of unaligned source words (USW) across the datasets (Bottom). Further, GPT translations obtain higher non-monotonicity scores for E-X translations (Middle).}
    \label{fig:measurements_figure}
    \vspace{-1.0em}
\end{figure}

\paragraph{Results} We compare the performance of the four systems on the WMT-21 test sets. Figure~\ref{fig:measurements_figure} shows the results of this comparison. A key observation is that while the GPT based translations achieve superior COMET-QE+LID scores than Microsoft Translator across the language pairs (except En-Ru), they also consistently obtain considerably higher number of unaligned source words. This result holds for the comparison between the WMT-21-SOTA and GPT systems as well. Further, GPT translations also consistently show higher non-monotonicity for E$\rightarrow$X translations. However, this is not the case for translations into English, wherein the multilingual WMT-21-SOTA system obtains very close non-monotonicity measurements. The \textit{combined interpretation} of these measurements \textit{suggests} that GPTs do produce less literal E$\rightarrow$X translations.

\paragraph{Human Evaluation} We verify the conclusion from the results in Figure~\ref{fig:measurements_figure} by conducting a human evaluation of translation literalness on 6 WMT-22 language pairs: En-De, En-Ru, En-Zh and De-En, Ru-En, Zh-En. 
For each language pair, we randomly sample 100 source-translation pairs, with translations obtained from MS-Translator (a strong commercial NMT system) and text-davinci-003 (a strong commercial LLM) \cite{mt-gpt}. 
We used zero-shot text-davinci-003 translations for human evaluations in order to eliminate any biases through the use of specific demonstration examples. 
In each case, we ask a human annotator (bilingual speaker for Zh-En, target-language native plus bilingual speaker otherwise) to annotate 100 translations from both GPT and MS-Translator and select which of the two translations is more literal. 
The human annotation interface is described in Appendix~\ref{sec:appendix_b}. 
The results in Table~\ref{tab:human} show that the annotators rate the GPT translations as less literal.

\begin{table}[!htbp]
\centering
\scalebox{0.75}{
\begin{tabular}{c|c|c|c|c}
\midrule
\textbf{Lang-Pair} & \textbf{MS-Translator} & \textbf{Davinci-003}  & \textbf{Equal} & \textbf{Diff} \\ \midrule
En-De & 52 & 32 & 16 & +20 \\
En-Zh & 42 & 32 & 24 & +10 \\
En-Ru & 41 & 37 & 22 & + 4 \\ \midrule
De-En & 48 & 26 & 26 & +12 \\
Zh-En & 42 & 38 & 20 & + 4 \\
Ru-En & 52 & 28 & 20 & +24 \\\midrule
\end{tabular}}
\caption{Human Evaluation Results across different language pairs on which is the \textit{more literal translation}: the numbers are from annotations done on 100 translations obtained from both MS-Translator and Davinci-003.}
\label{tab:human}
\vspace{-1.5em}
\end{table}

\paragraph{Experiments on Best WMT-22 NMT Systems} Further, we also experiment with the WMT-Best systems on the WMT-22 General Machine Translation task \cite{kocmi-etal-2022-findings}. We evaluate USW and NM on De-En, Ja-En, En-Zh and Zh-En, since on each of these language pairs, text-davinci-003's few-shot performance is very close to that of the WMT-Best system as per COMET-22 \cite{rei-etal-2022-comet}, based on the evaluation done in \citet{mt-gpt}. We report our results in Table~\ref{tab:wmt22}, which shows our prior findings replicated across the language pairs. For example, text-davinci-003, despite obtaining a 0.2 to 0.6 \textit{higher} COMET-22 score than the best WMT systems on these language pairs, consistently obtains a \textit{higher} USW score and a higher NM score in all but one comparison (NM for En-De). Note that the NM score differences for Chinese and Japanese are larger in magnitude owing to alignment deviations measured over character-level alignments. Further, we refer the reader to \citet{mt-gpt} for similar USW and NM comparisons of translations from text-davinci-003 and MS-Translator.

\begin{table}[!htbp]
\centering
\scalebox{0.99}{
\begin{tabular}{c|c|c}
\midrule
\textbf{Language Pair} & \textbf{USW Diff} & \textbf{NM Diff} \\ \midrule
En-Zh  & + 4.93 & + 12.94  \\ \midrule
De-En  & + 1.04 & - 0.10  \\ 
Zh-En  & + 4.93 & + 13.06  \\ 
Ja-En  & + 6.10 & + 11.13 \\ \midrule
\end{tabular}}
\caption{USW and NM score differences of text-davinci-003 relative to WMT-Best on the WMT-22 test sets.}
\label{tab:wmt22}
\vspace{-0.5em}
\end{table}

\section{Effects On Figurative Compositionality}
\label{sec:idioms}

In this section, we explore whether the less literal nature of E$\rightarrow$X translations produced by GPT models could be leveraged to generate higher quality translations for certain inputs. We posit the phenomenon of composing the non-compositional meanings of idioms \cite{dankers_compositionality} with the meanings of the compositional constituents within a sentence as \emph{figurative compositionality}. Thereby, a model exhibiting greater figurative compositionality would be able to abstract the meaning of the idiomatic expression in the source sentence and express it in the target language non-literally, either through a non-literal (paraphrased) expression of the idiom's meaning or through an equivalent idiom in the target language. Note that greater non-literalness does not imply better figurative compositionality. Non-literalness in a translation could potentially be generated by variations in translation different from the \textit{desired} figurative translation. 

\subsection{Translation with Idiomatic Datasets}

In this section, we quantify the differences in the translation of sentences with idioms between traditional NMT systems and a GPT model. There do not exist any English-centric parallel corpora dedicated to sentences with idioms. Therefore, we experiment with monolingual (English) sentences with idioms. The translations are generated with the same prompt in Section~\ref{sec:measures}. The datasets with \textit{natural idiomatic sentences} are enumerated below:

\begin{table}[!t]
\centering
\scalebox{0.99}{
\begin{tabular}{c|c|c|c}
\midrule
MT System & \textbf{C-QE $\uparrow$} & \textbf{USW $\downarrow$} & \textbf{NM $\downarrow$}\\ \midrule
MS-Translator   & 21.46 & 13.70 & 9.63 \\ 
WMT'21 SOTA  & 23.25 & 14.47 &  10.21 \\ 
text-davinci-002  & \textbf{23.67}  & \textbf{18.08} & \textbf{11.39} \\ \midrule
\end{tabular}}
\caption{Natural Idiomatic Sentences: Combined Results over MAGPIE, EPIE, PIE (5,712 sentences).}
\label{tab:encs}
\vspace{-1.0em}
\end{table}

\begin{itemize}
\item \textit{MAGPIE} \cite{magpie} contains a set of sentences annotated with their idiomaticity, alongside a confidence score. We use the sentences pertaining to the news domain which are marked as idiomatic with cent percent annotator confidence (totalling 3,666 sentences).
\item \textit{EPIE} \cite{epie} contains idioms and example sentences demonstrating their usage. We use the sentences available for static idioms (totalling 1,046 sentences).
\item The \textit{PIE dataset} \cite{zhou-etal-2021-pie} contains idioms along with their usage. We randomly sample 1K sentences from the corpus.
\end{itemize}
\paragraph{Results} The results are presented in Table~\ref{tab:encs}. We find that text-davinci-002 produces better quality translations than the WMT'21 SOTA system, with greater number of unaligned words as well as with higher non-monotonicity. 

\paragraph{Further Analysis} Note that a direct attribution of the gain in translation quality to better translation of idioms specifically is challenging. Further, similarity-based quality metrics such as COMET-QE themselves might be penalizing non-literalness, even though they are less likely to do this than surface-level metrics such as BLEU or ChrF \cite{bleu, chrf}. Therefore, while a natural monolingual dataset presents a useful testbed for investigating figurative compositionality abilities, an explicit comparison of figurative compositionality between the systems is very difficult. Therefore, we also conduct experiments on synthetic data, where we explicitly control the fine-grained attributes of the input sentences. We do this by allocating most of the variation among the input sentences to certain constituent expressions in synthetic data generation.

\subsection{Synthetic Experiments} 

For our next experiments, we generate synthetic English sentences, each containing expressions of specific \textit{type(s)}: (i) names, (ii) random descriptive phrases, and (iii) idioms.
We prompt text-davinci-002 in a zero-shot manner, asking it to generate a sentence with different \textit{instantiations} of each of these types (details are in appendix \ref{sec:appendix_c}).
We then translate these sentences using the different systems, in order to investigate the relative effects on our literalness metrics between systems and across types.
In each of the control experiments, we translate the synthetic English sentences to German.

\paragraph{Synthetic Dataset 1} As described, we generate sentences containing expressions of the three types, namely, named entities (e.g., \textit{Jessica Alba}), random descriptive phrases (e.g., \textit{large cake on plate}) and idioms (e.g., \textit{a shot in the dark}). 
Expression sources as well as further data generation details are presented in Appendix~\ref{sec:appendix_c}. 
Results are in Table~\ref{tab:idioms}.

\begin{table}[!t]
\centering
\scalebox{0.99}{
\begin{tabular}{c|c|c|c}
\hline
Expression & \textbf{C-QE $\uparrow$} & \textbf{USW $\downarrow$} & \textbf{NM $\downarrow$}\\ \hline
Random Phrases   &  -2.45  & +1.62 & +0.14 \\ 
Named Entities   & -1.50  & +0.81 & +0.39 \\ 
Idioms   & \textbf{+5.90} & \textbf{+2.82} & \textbf{+1.95} \\ \hline
\end{tabular}}
\caption{Synthetic sentences with Idioms vs Synthetic sentences containing other expressions: The difference between GPT (text-davinci-002) performance and NMT performance (Microsoft Translator) is reported.}
\label{tab:idioms}
\vspace{-1.5em}
\end{table}

\begin{table}[!htbp]
\centering
\scalebox{0.95}{
\begin{tabular}{c|c|c|c|c}
\hline
Num Idioms & \textbf{1} & \textbf{2} & \textbf{3} & \textbf{4} \\ \hline
\textbf{USW}   & 17.58 & 18.39 & 18.28 & 18.99 \\ \hline
\end{tabular}}
\caption{Synthetic sentences with multiple idioms (1-4): Increasing the number of idioms increases the number of unaligned source words in text-davinci-002 translations.}
\label{tab:synthetic}
\vspace{-1.2em}
\end{table}

\paragraph{Synthetic Dataset 2} We generate sentences containing \textit{multiple} idioms (varying from 1 to 4). The prompts \& examples are presented in appendix \ref{sec:appendix_c}. The results are presented in Table~\ref{tab:synthetic}.

\paragraph{Results} Table \ref{tab:idioms} shows that the percentage of unaligned source words is highest in the case of idioms, followed by random descriptive phrases \& named entities. The results are consistent with the hypothesis that the explored GPT models produce less literal E$\rightarrow$X translations, since named entities or descriptive phrases in a sentence would admit more literal translations as acceptable, unlike sentences with idioms. Davinci-002 obtains a much higher COMET-QE score in the case of translations of sentences with idioms, yet obtains a higher percentage of unaligned source words.
Similarly, the difference in non-monotonicity scores is also considerably higher for the case of idioms. 
These results provide some evidence that the improved results of the GPT model, together with the \textit{lower literalness} numbers, stem from correct translation of idiomatic expressions.
Table~\ref{tab:synthetic} shows that this effect only increases with the number of idioms.
\section{Discussion}
\label{sec:discussion}

In our experiments conducted across different NMT systems and GPT models, we find evidence that GPTs produce translations with greater non-literalness for E$\rightarrow$X in general.
There could be a number of potential causes for this; we list two plausible hypotheses below:

\paragraph{Parallel Data Bias} NMT models are trained on parallel data, which often contains very literal web-collected outputs.
Some of this may even be the output of previous-generation MT systems, which is highly adopted and hard to detect.
In addition, even high quality target text in parallel data always contains artifacts that distinguishes it from text originally written in that language, i.e. the `translationese' effect \cite{gellerstam}. These factors could likely contribute to making NMT translations comparatively more literal.

\paragraph{Language Modeling Bias} Translation capability in GPTs arises in the absence of any \textit{explicit} supervision for the task during the pre-training stage. Therefore, the computational mechanism that GPTs leverage for producing translations might be different from NMT models, imparting them greater abstractive abilities. This could have some measurable manifestation in the translations produced, e.g., in the literalness of the translations.

\paragraph{Differences in E$\rightarrow$X and X$\rightarrow$E} In E$\rightarrow$X, we consistently find that GPT translations of similar quality are less literal and in the X$\rightarrow$E direction, we observe a few anomalies. For X$\rightarrow$E, in Figure~\ref{fig:measurements_figure}, in all but one comparison (WMT-21-SOTA vs GPTs for De-En) GPTs obtain higher measures for non-literalness. On the other hand, we did not see anomalies in the trend for E$\rightarrow$X directions.

\paragraph{Variations in Experimental Setup} We also experimented with a variant of USW and NM which doesn't use the alignments pertaining to stopwords. Each of our findings remain the same, with relatively minor changes in magnitudes but not in system rankings. Similarly, we observed a greater tendency towards less literalness in GPT translations in both few-shot and zero-shot settings, when compared across a range of NMT systems.
  
\section{Summary and Conclusion}
\label{sec:summary}

We investigated how the translations obtained through LLMs from the GPT family are qualitatively different by quantifying the property of translation literalness. We find that for E$\rightarrow$X translations, there is a greater tendency towards non-literalness in GPT translations. 
In particular, this tendency becomes evident in GPT systems' ability to figuratively translate idioms.

\section{Acknowledgements}
\label{sec:ack}

We thank Hitokazu Matsushita for help in conducting human evaluations.

\section{Limitations}
Measurement of translation literalness is neither well studied nor well understood. 
We rely on a combined interpretation of multiple measurements to investigate our hypothesis and its implications. This limits the extent to which we can make strong claims, since in the absence of a highly correlated metric for translation literalness, it is hard to compare systems. 
We could only claim that our investigation indicates the presence of a tendency towards non-literalness in GPT translations, but a stronger result would have been preferred to further disambiguate the translation characteristics. Further, we only compare GPT translations in the standard zero-shot and few-shot settings and it is quite conceivable that more specific \& verbose instructions could steer the LLMs to produce translations with different characteristics.

\bibliography{anthology,custom}
\bibliographystyle{acl_natbib}

\appendix

\begin{figure}[ht]
\centering
\begin{subfigure}[b]{0.49\textwidth}
\centering
\includegraphics[width=\textwidth]{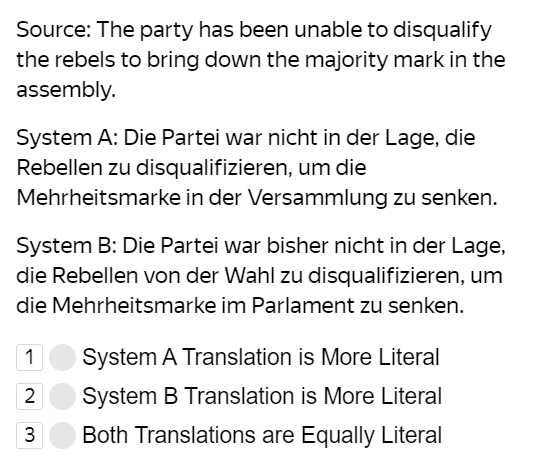}
\end{subfigure}
\caption{Annotation Interface Screenshot for Human Evaluation. The translations are randomized b/w the systems under evaluation to prevent biased evaluation.}
\label{fig:interface}
\vspace{-0.1em}
\end{figure}

\section{Human Annotation Interface}
\label{sec:appendix_b}

We use the annotation interface in Figure \ref{fig:interface}, in which the annotators are asked to rate the two translations. The bilingual and native speaker annotators were recruited in-house.

\section{Synthetic Dataset Details}
\label{sec:appendix_c}
\textbf{Synthetic Dataset 1}: For each of the three expression types, 100 synthetic sentences are generated. Figures \ref{fig:prompt1}, \ref{fig:prompt2}, \ref{fig:prompt3} present examples. The sources of the named entities and descriptive phrase expressions are MultiNERD \cite{multinerd} and PhraseCut \cite{phrasecut} datasets respectively. 

\begin{figure}[tbh]%
\centering%

\fbox{\parbox{\dimexpr\linewidth-2\fboxsep+2.5\fboxrule\relax}{ 
\textbf{Prompt}: Q: Generate a sentence containing the idiom: \textbf{a short fuse}, in the form of a news article sentence. \textbackslash n A:  \\
\textbf{Output}: \textit{The man was known to have \textbf{a short fuse}, and often exploded into fits of anger without warning.}
}}%
\caption[]{\textbf{Synthetic Dataset 1}: Example (\textbf{Idiom})}%

\label{fig:prompt1}%
\end{figure}

\begin{figure}[tbh]%
\centering%

\fbox{\parbox{\dimexpr\linewidth-2\fboxsep+2.5\fboxrule\relax}{ 
\textbf{Prompt}: Q: Generate a sentence containing the entity: \textbf{Wolfgang Amadeus Mozart}, in the form of a news article sentence. \textbackslash n A:  \\
\textbf{Output}: \textit{\textbf{Wolfgang Amadeus Mozart}, a world renowned composer, died today at the age of 35.}
}}%
\caption[]{\textbf{Synthetic Dataset 1}: Example (\textbf{Entity})}%

\label{fig:prompt2}%
\end{figure}

\paragraph{Synthetic Dataset 2:} Figure \ref{fig:prompt4} presents an example of the prompt as well as for a generated synthetic sentence containing two idioms.

\begin{figure}[tbh]%
\centering%
\fbox{\parbox{\dimexpr\linewidth-2\fboxsep+2.5\fboxrule\relax}{ 
\textbf{Prompt}: Q: Generate a sentence containing the phrase: \textbf{white chair}, in the form of a news article sentence. \textbackslash n A:  \\
\textbf{Output}: \textit{The \textbf{white chair} was found to be comfortable by the majority of the participants in the study.}
}}%
\caption[]{\textbf{Synthetic Dataset 1}: Example (\textbf{Phrase})}%
\label{fig:prompt3}%
\end{figure}

\begin{figure}[tbh]%
\centering%
\fbox{\parbox{\dimexpr\linewidth-2\fboxsep+2.5\fboxrule\relax}{ 
\textbf{Prompt}: Q: Generate a sentence using the two idioms: \textbf{off the wall}, \textbf{claim to fame} in the form of a news article sentence. \textbackslash n A:  \\
\textbf{Output}: \textit{The company's \textbf{off-the-wall} marketing campaign was its \textbf{claim to fame}.}
}}%
\caption[]{\textbf{Synthetic Dataset 2}: Example (2 \textbf{Idioms})}%
\label{fig:prompt4}%
\end{figure}

\section{Alignments and Literalness}
\citet{literalness_measuring} find that more alignment crossings (which is measured by the non-monotonicity metric) between the source and translations are proportional to the extra cognitive effort (measured using gazing time of human translators) required by human translators in processing non-literal translations. This links alignment crossings (the non-monotonicity measure is normalized alignment crossing) with greater non-literalness.  

\end{document}